\documentclass[spdflatex,n-nature]{sn-jnl}

\usepackage{graphicx}%
\usepackage{multirow}%
\usepackage{amsmath,amssymb,amsfonts}%
\usepackage{amsthm}%
\usepackage{mathrsfs}%
\usepackage[title]{appendix}%
\usepackage{xcolor}%
\usepackage{textcomp}%
\usepackage{manyfoot}%
\usepackage{booktabs}%
\usepackage{algorithm}%
\usepackage{algorithmicx}%
\usepackage{algpseudocode}%
\usepackage{listings}%

\theoremstyle{thmstyleone}%
\theoremstyle{thmstyletwo}%

\theoremstyle{thmstylethree}%

\newcommand{\emrank}{\textsc{3EM-Ranker}}

\usepackage{pifont}
\usepackage{booktabs}   
\usepackage{multirow}   
\usepackage{makecell}

\usepackage[most]{tcolorbox}

\usepackage[most]{tcolorbox}
\usepackage{caption}
\usepackage{threeparttable}
\newtcolorbox{promptbox}{
  colback=gray!5,
  colframe=gray!60,
  boxrule=0.5pt,
  arc=2mm,
  left=6pt, right=6pt, top=6pt, bottom=6pt
}

\begin{document}

\title[Article Title]{From Generation to Collaboration:
Using LLMs to Edit for Empathy in Healthcare}


\author*[1]{\fnm{Man} \sur{Luo}}\email{luoman.cs@gmail.com}
\equalcont{These authors contributed equally to this work.}

\author[2]{\fnm{Bahareh} \sur{Harandizadeh}}\email{Harandizadeh.Bahareh@mayo.edu}
\equalcont{These authors contributed equally to this work.}

\author[3]{\fnm{Amara} \sur{Tariq}}\email{Tariq.Amara@mayo.edu}

\author[2]{\fnm{Halim} \sur{Abbas}}\email{Abbas.Halim@mayo.edu}

\author[4]{\fnm{Umar} \sur{Ghaffar}}\email{Ghaffar.Umar@mayo.edu}

\author[4]{\fnm{Christopher} \sur{J Warren}}\email{warren.christopher@mayo.edu}

\author[2]{\fnm{Segun O.} \sur{Kolade}}\email{Kolade.John@mayo.edu}

\author[4]{\fnm{Haidar M.} \sur{Abdul-Muhsin}}\email{Abdul-Muhsin.Haidar@mayo.edu}

\affil*[1]{\orgdiv{Science team}, \orgname{Abridge}, \orgaddress{\street{470 Alabama}, \city{San Francisco}, \postcode{94110}, \state{CA}, \country{U.S}}}

\affil[2]{\orgdiv{Platform}, \orgname{Mayo Clinic}, \orgaddress{\street{200 First St. S.W.}, \city{Rochester}, \postcode{10587}, \state{MN}, \country{U.S}}}

\affil[3]{\orgdiv{Department of AI \& Informatics}, \orgname{Mayo Clinic}, \orgaddress{\street{6161 E Mayo Blvd}, \city{Phoenix}, \postcode{85054}, \state{AZ}, \country{U.S}}}

\affil[4]{\orgdiv{Urology Department}, \orgname{Mayo Clinic}, \orgaddress{\street{5777 E Mayo Blvd}, \city{Phoenix}, \postcode{85054}, \state{AZ}, \country{U.S}}}


\abstract{
Clinical empathy is essential for patient care, but physicians need continually balance emotional warmth with factual precision under the cognitive and emotional constraints of clinical practice. This study investigates how large language models (LLMs) can function as empathy editors, refining physicians’ written responses to enhance empathetic tone while preserving  underlying medical information. More importantly, we introduce novel quantitative  metrics, an Empathy Ranking Score and a MedFactChecking Score to systematically assess both emotional and factual quality of the responses. Experimental results show that LLM-edited responses significantly increase perceived empathy while preserving factual accuracy compared with fully LLM-generated outputs. These findings suggest that using LLMs as editorial assistants, rather than autonomous generators, offers a safer, more effective pathway to empathetic and trustworthy AI-assisted healthcare communication.}




\maketitle

\section{Introduction}\label{intro}
Empathy encompasses cognitive understanding of others' emotional states and affective emotional responses when witnessing others' emotions \cite{cuff2016empathy, davis1983measuring}. In healthcare, clinical empathy involves understanding patient pain and suffering, communicating this understanding, and having an intention to help \cite{hojat2007empathy}. Clinical empathy plays a crucial role in patient care, as it fosters trust, encourages patient engagement, and ultimately leads to better treatment outcomes \cite{kim2004effects, hojat2011physicians, rakel2011perception}. Studies have shown that empathetic communication improves adherence to medical advice and enhances patient satisfaction. Despite its importance, empathy is often difficult to sustain in routine clinical practice \cite{derksen2012effectiveness, kelley2014influence}. Physicians face heavy cognitive and emotional demands, including time pressure, complex decision making, and repeated exposure to distressing situations. To safeguard their own mental well-being and maintain sound clinical judgment, they must balance empathy with a degree of professional detachment \cite{larson2005clinical, gleichgerrcht2013empathy, neumann2011empathy}. This creates a persistent tension between factual precision and emotional warmth, posing a key challenge in preserving clinical  accuracy and objectivity while ensuring compassionate  physician–patient communication. 

\begin{figure}[t]
    \centering
    \includegraphics[width=0.95\textwidth]{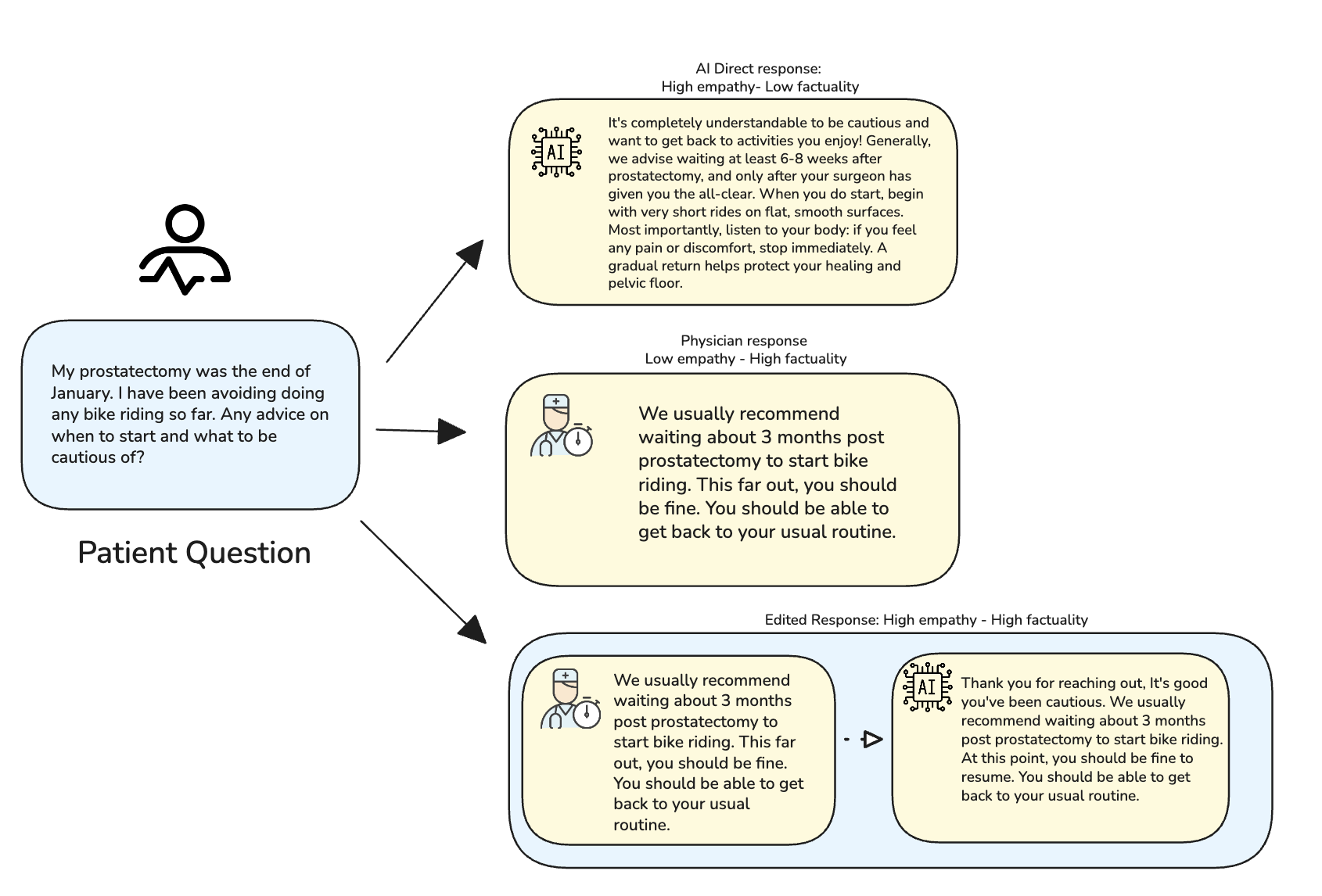}
    \caption{AI editing versus generation for clinical responses. Direct AI generation produces empathetic but factually inaccurate responses with hallucinated medical details, while AI editing of physician responses maintains factual accuracy while adding empathy.}
    \label{fig:fig_1}
\end{figure}

Recent advances in large language models (LLMs) have revolutionized healthcare applications \cite{thirunavukarasu2023large, lee2023benefits}. Clinical chatbots and decision support tools have shown potential to improve accessibility for patients and reduce the workload of healthcare professionals \cite{ayers2023comparing}. However, their deployment in clinical settings raises important concerns regarding patient privacy \cite{wang2023ethical, mello1938chatgpt}, data security, and the factual accuracy of AI-generated content \cite{singhal2023large, omiye2023large}. Consequently, most prior studies have focused primarily on clinical accuracy and safety,  often at the expense of empathetic communication. Moreover, many current systems rely on end-to-end text generation, which increases the risk of hallucination and reduces control over tone, intent, and emotional expression \cite{ji2023survey, huang2025survey}.

To address the lack of empathy in AI-generated medical responses, researchers have explored fine-tuning LLMs on empathetic medical dialogues and patient-physician conversations \cite{xu2024revolutionizing, yosef2025impact, welivita2020taxonomy}. While fine-tuning can enhance empathetic language generation, it requires substantial computational resources and domain expertise, making it costly and time-intensive \cite{wornow2023shaky}. More critically, studies have shown that fine-tuning LLMs for stylistic attributes such as empathy can inadvertently increase the risk of factual errors, fabrications, and ``phony empathy'' \cite{ibrahim2025training}. In the medical domain, where factual precision is paramount, such trade-offs pose significant safety concerns. These limitations highlight the need for alternative approaches that can enhance empathy without compromising clinical accuracy or requiring extensive model retraining.

To address these limitations, we propose reframing the role of LLMs in clinical communication: from autonomous generators to collaborative editors that refine physician responses to enhance empathy while maintaining factual integrity. This perspective emphasizes augmentation rather than automation, positioning LLMs as partners that enhance, rather than replace, human expertise. Figure~\ref{fig:fig_1} illustrates our proposed collaborative editing approach. In this example, a patient asks about resuming bike riding after a prostatectomy, the physician provides an accurate but terse response. An end-to-end AI chat system produces a more empathetic message but introduces factual errors. Our editing approach preserves the physician's  recommendation while adding empathetic acknowledgment. This demonstrates how LLMs can enhance physician-patient communication without introducing hallucinations or compromising medical precision.

To systematically evaluate this editing approach, we explore three research questions: 
\begin{itemize}
    \item \textbf{RQ1:} How can we add empathy to physician responses and measure the improvement?
    \item \textbf{RQ2:} How can we ensure empathetic edits preserve clinical facts?
    \item \textbf{RQ3:} How do empathy and factuality relate to each other?
\end{itemize}

We developed two complementary evaluation protocols. For i) \emph{empathy evaluation}, we introduce a three-way empathy comparison which achieves decent alignment score with human evaluators. For ii) \emph{factuality evaluation}, we introduce the MedFactChecking Score to measure Fact-Recall (preservation of original medical information) and Fact-Precision (grounding of enhanced content in the original response). Through bidirectional entailment checks, we detect both information loss and hallucinated additions. Domain experts validated both metrics and reviewed responses across multiple models and empathy levels, informing iterative prompt refinement to balance empathy with factual integrity.

Based on our investigations of RQ1–RQ3, our experiments demonstrate that the proposed empathy editing framework, equipped with a refined editing instruction prompt, achieves the most favorable trade-off between empathy and factual accuracy, making it well suited for practical clinical use. Overall, this work reframes LLMs as safe, empathetic, and trustworthy editorial assistants that support clinicians in producing emotionally resonant yet factually sound communication, which we consider as an essential step toward responsible AI integration in healthcare.



\section{Related Work}\label{related_work}

Clinical empathy improves trust, treatment adherence, and patient satisfaction \cite{hojat2011physicians, derksen2012effectiveness}, yet physicians must balance emotional warmth with clinical detachment to preserve factual precision under time pressure \cite{larson2005clinical, gleichgerrcht2013empathy}. This tension between empathy and accuracy remains underexplored in NLP research, which lacks concrete conceptualization of empathy in text \cite{shetty2024scoping, lahnala2022critical}. 

Central to this challenge is the question of how empathy should be measured. Empathy measurement in healthcare relies on validated psychometric instruments such as the Jefferson Scale of Empathy (JSE) \cite{hojat2018jefferson} and the Consultation and Relational Empathy (CARE) scale \cite{hemmerdinger2007systematic}. While these instruments reliably assess human empathy, they were designed for human interactions rather than algorithmic text generation, making their direct application to LLM-generated text problematic \cite{shteynberg2024does}.

To address this gap, researchers have developed computational approaches for empathy evaluation. These include rule-based frameworks, machine learning, and context grammar approaches \cite{dey2023investigating, dey2022enriching, provence2024algorithms}. For AI-generated text specifically, methods include RoBERTa-based frameworks \cite{sharma2020computational} and LLM-based approaches like EM-Rank \cite{luo2024assessing} and ESC-Rank \cite{zhao2024esc}. In-context learning has also been used to evaluate LLM therapists, finding they tend to mimic low-quality therapy \cite{chiu2024computational}. However, reliability and alignment with human judgment remain concerns for these automated methods \cite{sorin2024large}.

Beyond evaluation methods, a separate line of research has examined the inherent empathetic capabilities of LLMs. Studies show that while LLMs are generally perceived as empathetic, they exhibit notable limitations \cite{sorin2024large, huang2024apathetic}, including repetitive use of empathetic phrases \cite{huang2024apathetic} and lack of cultural understanding compared to human counselors \cite{iftikhar2024therapy}. Additionally, highly educated patients report greater skepticism toward chatbot performance \cite{carl2025patient}, suggesting that perceived empathy may vary across user populations.

Given these limitations, researchers have developed various approaches to enhance empathy in LLM-generated text. These include augmenting LLMs with small-scale empathetic models \cite{yang2024enhancing}, integrating psychotherapy and psychological principles into LLM frameworks \cite{lee2024enhancing, wang2024empllm}, and leveraging human-AI collaboration to improve empathic conversations \cite{sharma2023human}. LLM-generated feedback has also been shown to improve empathy expression among healthcare trainees \cite{yao2024enhancing}. However, fine-tuning LLMs for empathy requires substantial computational resources and extensive labeled datasets \cite{wiggins2022opportunities}, posing practical limitations for widespread implementation.

While these enhancement efforts focus on improving empathy, they rarely consider potential trade-offs with factual accuracy. There is growing concern regarding the balance between empathy enhancement and maintaining factuality in AI-generated text \cite{ibrahim2025training}, especially for sensitive domains like medicine and legal proceedings \cite{ongai, zhanghappens, zhang2024empathetic}. However, systematic investigation of their relationship in clinical contexts remains limited, particularly regarding which specific medical information is vulnerable to degradation during empathy enhancement.

\section{Methods} \label{method}
\subsection{Empathy Edition}\label{empathy_edition}
Previous work has explored using LLMs to generate an empathetic responses by just giving the patient question. While these LLM-generated responses tend to be more empathetic than those written by physicians~\cite{luo2024assessing}, this approach presents two major challenges in real-world applications. First, there are privacy concerns, as patient data must be carefully handled to prevent unintended disclosure. Prior studies have addressed this by performing manual de-identification, which is labor-intensive. Second, there is a factuality concern, as LLMs can generate responses that contain factual errors, potentially leading to misinformation in medical communication.

To mitigate these issues, we propose an alternative approach: instead of generating responses purling by LLMs, we use LLMs to edit physician responses. This ensures that the editing responses are grounded by factual contents while enhancing empathetic expression. Our method positions LLMs as controlled editors, focusing on linguistic empathy rather than medical inference, thereby minimizing the risk of hallucinated or speculative statements.

\paragraph{Simple Editing Prompt}
\label{Simple Editing Prompt}
We design an instruction prompt that explicitly directs the model to revise 
physician responses to enhance empathy while maintaining the original meaning 
and writing style (see Appendix~\ref{app:prompts:empathy} for full prompt templates). 
The prompt provides the patient's question and the physician's response, 
ensuring the model has sufficient context for effective editing. Additionally, 
we experiment with a minimal-editing prompt, which instructs the model to make 
only essential modifications, preserving as much of the original response as 
possible. This setup allows us to isolate the linguistic effects of empathy 
editing from those of content generation.

\paragraph{Refine Editing Prompt}
Using the simple editing prompt, we obtained model outputs and engaged a 
domain expert to evaluate the quality of these responses. The expert received 
all pairs of responses, consisting of the original patient message, the 
original physician response, and the model-edited version. The initially 
generated editing responses occasionally included factual inaccuracies and 
speculative statements (more details are discussed in~\ref{sec:MedFactChecking Validation}). 
To mitigate these risks, we designed a refined prompt by adding explicit 
behavioral constraints to the simple editing prompt (see Appendix~\ref{app:prompts:empathy}). 
These editing principles define the operational boundaries for empathy 
enhancement, ensuring that model interventions remain clinically grounded 
and ethically appropriate. Throughout our experiments in \S\ref{exp_analysis}, 
we employ both simple and refined prompts and compare the resulting levels 
of empathy and factuality.

\subsection{Empathy Evaluation}\label{empathy_evaluation}
Most prior studies have assessed clinical empathy through human annotation~\cite{rashkin2019towards,hosseini2021takes,sharma2020computational}, while a recent work, EMRanknk~\cite{luo2024assessing} has introduced automated evaluation using LLMs as clinical empathy raters. Although these LLM-based assessments show moderate agreement with human judgments, their reliability remains limited. In this work, we develop a new empathy metric based on LLMs-judge to align better with human preference. 
In this section, we provide an overview of EMRank and our proposed 3-EMRank, which achieves a higher degree of alignment with human judgments.


\noindent\textbf{Overview of the  EMRank framework.} EMRank is a prompt-based evaluation method inspired by the LLMs-as-judge paradigm~\cite{zheng2023judging}. It utilizes a large language model (LLM) as a judge to determine which response is more empathetic. The evaluation is conducted in zero-shot, one-shot, and few-shot settings, with an ensemble approach that aggregates majority votes from these settings. Among these methods, the ensemble approach has been shown to achieve the highest alignment with human judgments. For more details, we refer the readers to the original paper. A key limitation of the original EMRanknk framework is that it relies on binary comparison, forcing the model to always choose one response as more empathetic than the other. However, in real-world scenarios, two responses may exhibit similar levels of empathy, making binary comparisons inadequate and potentially contributing to discrepancies between human and model evaluations.


\noindent\textbf{Implementation of Three-way EMRank.} 
To address the issue of EMRank, we introduce a three-way ranking, \emrank{}, which allows for an additional judgment option: both responses are equally empathetic. This modification enables more nuanced differentiation among comparable responses and better reflects the subjectivity inherent in empathy perception. The three-way extension is achieved by modifying the evaluation prompt to include a third response category, ``Both responses are equally empathetic." This adjustment improves the flexibility of the model judgment and improves the consistency with human ratings. We also test multiple LLMs as empathy judges to ensure the cross-model reliability of the EMRank outputs.

The modified prompts shown in Supplementary Fig.~\ref{fig:3emrank_prompt} in Appendix~\ref{app:prompts:Evaluation} explicitly accommodate equivalence judgments, enabling the metric to capture subtle distinctions between near-identical empathetic tones. 

\subsection{Factuality Evaluation} 
\label{factuality_evaluation}

\begin{figure}[t]
    \centering
    \includegraphics[width=1.0\textwidth]{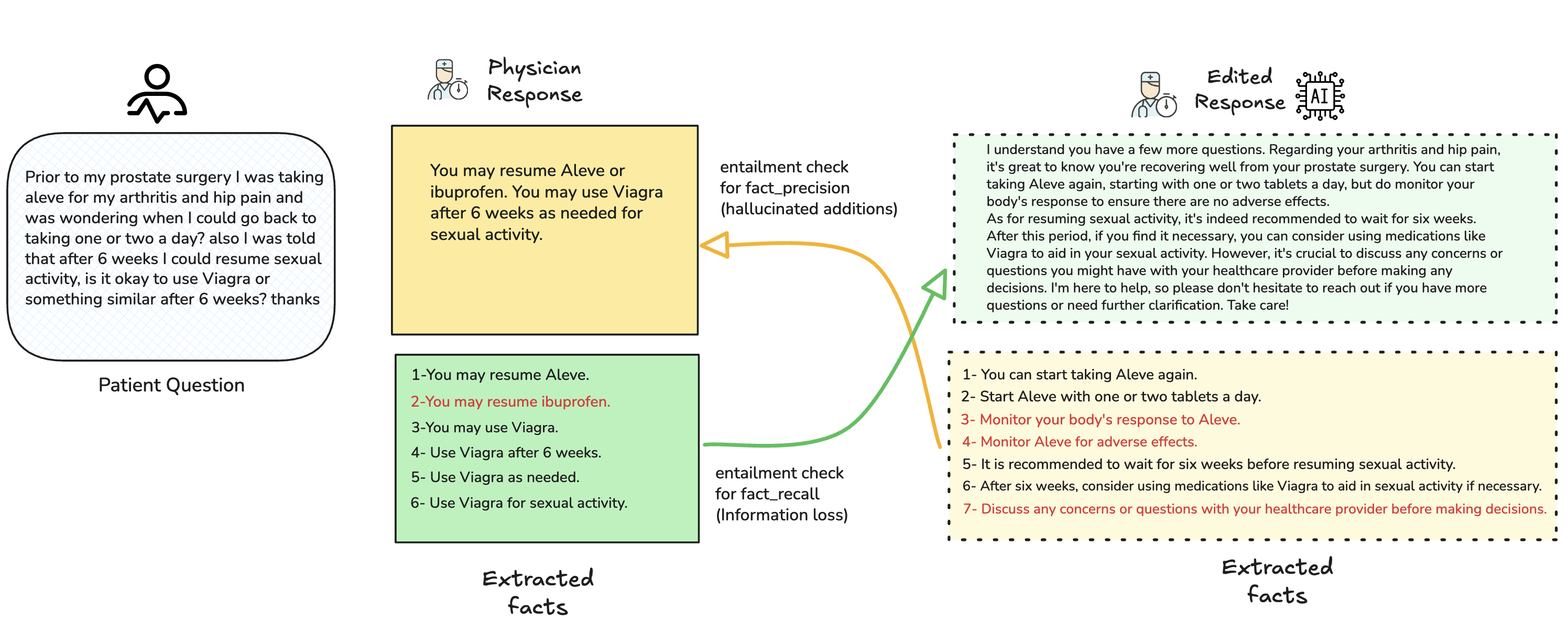}
    \caption{Bidirectional fact-checking framework for measuring factual accuracy in AI-edited clinical responses. Fact-Recall (green arrow) quantifies information loss from the original physician response, while Fact-Precision (yellow arrow) detects hallucinated additions in the edited response. Red text indicates facts that fail entailment checking: lost information (left) and unsupported additions (right).}
    \label{fig:fig_2}
\end{figure}

To systematically evaluate whether empathy enhancement preserves medical accuracy, we introduce the MedFactChecking Score, an automated evaluation framework adapted from FactEHR~\cite{munnangi2025factehr}. Unlike FactEHR, which evaluates factuality within a single clinical note, our approach performs \textit{bidirectional} evaluation across two distinct texts: the original physician response and its empathy-enhanced version (Figure~\ref{fig:fig_2}). This bidirectional design is necessary because empathy enhancement can both lose information from the original response and introduce unsupported additions.

The framework quantifies two complementary dimensions: \textit{Fact-Recall} measures what proportion of medical facts from the original response are preserved in the enhanced version, while \textit{Fact-Precision} measures what proportion of facts in the enhanced response are grounded in the original. By extracting medical facts from both responses and performing entailment checks in both directions, we separately quantify omission and hallucination errors.

We validate our automated metrics through human annotation studies: ML experts verify the accuracy of fact extraction and entailment judgments, while two clinical experts (urologists) assesses whether MedFactChecking captures clinically significant errors and identifies patterns in what medical information is systematically lost or added.

\textbf{Formal Definition.} Let $d$ denote the original physician response and $d'$ denote the empathy-enhanced response. Let $C = \{c_1, c_2, \ldots, c_n\}$ represent the set of medical facts produced by decomposing $d$, and let $C' = \{c'_1, c'_2, \ldots, c'_m\}$ represent the set of medical facts produced by decomposing $d'$.

We define the entailment indicator function as:
\begin{itemize}
    \item $\mathbb[d' \models c] = 1$ if a fact $c \in C$ is completely entailed by the empathy-enhanced response $d'$, and 0 otherwise.
    \item $\mathbb[d \models c'] = 1$ if a fact $c' \in C'$ from the enhanced response is entailed by the original response $d$, and 0 otherwise.
\end{itemize}

Fact-Recall extracts facts from the original physician response and verifies their entailment in the enhanced response, thereby measuring information preservation. We define:
\begin{equation}
\text{Fact-Recall} = \frac{1}{|C|} \sum_{c \in C} \mathbb[d' \models c]
\label{eq:fact-recall}
\end{equation}

A higher recall indicates better preservation of the original medical information (i.e., less information loss during editing).
Fact-Precision extracts facts from the enhanced response and verifies their entailment in the original response, thereby detecting unsupported additions or hallucinations. We define:

\begin{equation}
\text{Fact-Precision} = \frac{1}{|C'|} \sum_{c' \in C'} \mathbb[d \models c']
\label{eq:fact-recall}
\end{equation}
A higher precision indicates fewer hallucinated medical facts added without basis in the original response. Together, these two measures provide complementary perspectives on factual reliability - recall captures retention, while precision captures restraint.

\textbf{Micro and Macro Averaging.} To provide a comprehensive evaluation, we report both averaging schemes. 
Macro-averaging computes metrics for each question-response pair independently, then averages across all pairs, treating each exchange equally regardless of fact count. Micro-averaging aggregates all fact counts across the corpus before computing metrics, giving more weight to responses with more facts. Formally:

\begin{align}
\text{Macro-Rec} &= \frac{1}{N} \sum_{i=1}^{N} \frac{\sum_{c \in C_i} \mathbb[d'_i \models c]}{|C_i|}, \quad
\text{Micro-Rec} = \frac{\sum_{i=1}^{N} \sum_{c \in C_i} \mathbb{I}[d'_i \models c]}{\sum_{i=1}^{N} |C_i|} \label{eq:recall}\\[1em]
\text{Macro-Prec} &= \frac{1}{N} \sum_{i=1}^{N} \frac{\sum_{c' \in C'_i} \mathbb[d_i \models c']}{|C'_i|}, \quad
\text{Micro-Prec} = \frac{\sum_{i=1}^{N} \sum_{c' \in C'_i} \mathbb[d_i \models c']}{\sum_{i=1}^{N} |C'_i|} \label{eq:precision}
\end{align}

where $N$ is the total number of question-response pairs. For edge cases where $C$ or $C'$ is empty, we define the metric as 1.0 if both sets are empty, and 0.0 otherwise. These conventions ensure metric stability even for brief or minimally factual responses.

\paragraph{Prompts for Fact Extraction and Entailment Checking}
\label{sec:prompts}

We employ two distinct prompts adapted from prior work: one for medical fact decomposition and another for entailment evaluation. This modular structure allows the same LLM to perform both extraction and verification tasks using clear, role-specific instructions.

\noindent\textbf{Fact Decomposition.} Our decomposition prompt, adapted from FActScore~\cite{min-etal-2023-factscore}, which is later also used by FactEHR~\cite{munnangi2025factehr}, includes 2 in-context examples and instructs the LLM to extract medical facts only, explicitly excluding empathetic expressions and conversational acknowledgments. For example, from ``I'm sorry you're worried. Your PSA is 4.2,'' we extract only ``PSA is 4.2''. This distinction is critical because empathy enhancement primarily affects emotional tone rather than medical content, and including non-medical statements would conflate empathy with factuality. The structured prompt shown in Supplementary Fig.~\ref{fig:fact-extraction-prompt} in Appendix~\ref{app:prompts:Evaluation}

\noindent\textbf{Entailment Evaluation Prompt.} For entailment verification, we use the prompt developed in FactEHR~\cite{munnangi2025factehr}, which was optimized using 40 premise-hypothesis pairs sampled from clinical notes and tuned to maximize F1 score on entailment classification. This prompt instructs the LLM to determine whether a given fact is completely entailed by, contradicted by, or neutral with respect to a source text, providing robust entailment judgments for medical statements. We maintain strict output formatting requirements to facilitate automated parsing of entailment predictions (exact prompt is given in Supplementary Fig.~\ref{fig:entailment-prompt} in Appendix~\ref{app:prompts:Evaluation}).

\subsection{QA Dataset Collection}
\label{dataset}

Our dataset comprises 163 patient-physician message pairs collected from a patient portal at a hospital. We focus on messages from individuals diagnosed with prostate cancer who underwent radical prostatectomy between December \textit{2018} and October \textit{2023}. Messages were gathered under the category of ``Patient advice request'' and randomly selected from the pool of available conversations (and we only use the first QA pair in the conversation). IRB approval was obtained to use the data.

\paragraph{De-identification Process} The de-identification process was meticulously carried out by a team of three: two medical students and one physician (postgraduate year 3 urology resident), with the primary goal of ensuring privacy and confidentiality. This process involved a detailed review of each patient message and physician response to identify and remove or anonymize any personally identifiable information (PII) including patient names, physician names, dates, phone numbers, and addresses.

\paragraph{Dataset Characteristics} 
Physician responses in our dataset are notably concise (mean 65.8±50.4 words, 
4.7±3.0 sentences), reflecting the time-constrained nature of asynchronous 
clinical communication in patient portals. Patient questions tend to be longer 
and more variable (mean 82.2±56.2 words, 6.4±4.1 sentences), with some patients 
providing extensive clinical context while others pose brief queries 
(see Supplementary Fig.~\ref{fig:dataset_stats} for full distributions). 
This heterogeneity in response length motivates our investigation of how 
brevity impacts factual preservation during empathy enhancement 
(see \S\ref{exp_analysis}). 

We also analyzed whether questions could be answered using only general 
medical knowledge or required patient-specific information from electronic 
health records (EHR). Using Gemini-2.5-Flash for classification, we found 
that 91.4\% (149/163) of questions require access to patient-specific EHR 
data (e.g., lab results, imaging findings, treatment history), while only 
8.6\% (14/163) represent general medical inquiries. This finding indicates 
that the task is largely evidence-dependent and simply using an end-to-end 
QA model without retrieving the correct EHR data is insufficient. In contrast, 
our proposed approach of editing existing physician responses offers a more 
realistic and clinically grounded framework.

\subsection{Models}
\label{Models}
We evaluate five state-of-the-art LLMs spanning both open-source and proprietary architectures: Qwen3-7/14B~\cite{yang2025qwen3}, Mistral-7B-Instruct~\cite{jiang2023mistral7b}, Llama-3.1-8B-Instruct~\cite{dubey2024llama}, Gemini-2.5-Flash~\cite{comanici2025gemini}. All models are hosted internally on the Mayo Clinic Platform within HIPAA-compliant compute environments to ensure patient data privacy and security.
We use four models for empathy-based editing and employ Llama-3.1 and Qwen3 for EM-Ranker evaluation. For fact decomposition and entailment verification—the core components of factuality evaluation, we use Gemini-2.5-Flash exclusively, following FactEHR's findings that Gemini achieves superior precision and recall on these tasks compared to alternative models.
\subsection{Metric Validation}
\paragraph{\emrank{} Validation}
To validate our three-way annotation approach, we conducted a human evaluation 
as a baseline for comparison against \emrank{}. We provided annotators with a 
patient question and two responses: one from a physician and one from an AI 
model (Gemini-2.5-Flash). Annotators selected from three options: response 1 
is more empathetic, response 2 is more empathetic, or they are equally 
empathetic. The patient annotators included three male patients with prostate 
cancer who had undergone radical prostatectomy within the same timeframe as 
our collected patient message dataset.

We compared the results produced by human annotators and \emrank{} 
(see Supplementary Table~\ref{tab:alignment-scores} for full results). 
When using Qwen-3 and LLaMA-3 as the underlying LLMs for \emrank{}, 
the alignment scores reach 0.57 and 0.55 respectively, both substantially 
higher than the previously reported 0.23 agreement between humans and 
EM-Rank \cite{luo2024assessing}. These results demonstrate that \emrank{} 
achieves considerably stronger alignment with human judgments compared to 
prior work.

\paragraph{MedFactChecking Validation}
\label{sec:MedFactChecking Validation}

To validate the MedFactChecking Score's ability to detect factual preservation 
in empathy-enhanced responses, we conducted a multi-stage evaluation combining 
automated fact extraction with expert human annotation. We edited physician 
responses using the Simple Editing prompt (see Appendix~\ref{app:prompts:empathy} or \S\ref{Simple Editing Prompt}) 
with Llama-3.1 to enhance empathy, then analyzed the responses using the 
MedFactChecking Score framework, ML researchers, and a medical expert.

\noindent\textbf{Automated Fact Extraction Validation.}
Two machine learning scientists independently verified the accuracy of facts 
flagged as added (hallucinated, $[d \models c'] = 0$) or not preserved (missed, 
$[d' \models c] = 0$) by the algorithm. For original responses, 934 facts were 
extracted, of which 191 were flagged as ``not preserved'' in the edited 
versions; annotators confirmed 139 of these (72.7\%) as genuinely 
missing. For edited responses, 1230 facts were extracted, of which 262 were 
flagged as ``added''; annotators confirmed 219 of these (83.6\%) as 
actual hallucinations (see Supplementary Table~\ref{tab:extraction_validation}).

The moderate results, particularly for not-preserved facts (72.6\%), warrant further interpretation. Many extraction errors involved semantically similar facts where the algorithm made overly conservative entailment judgments. While the semantic content was largely preserved between original and edited responses, some variations in phrasing led the algorithm to flag facts as ``not preserved'' or ``added.'' Annotators judged these conservative flaggings as incorrect since the core meaning remained intact. However, such errors do not necessarily indicate that the algorithm misses clinically meaningful changes. To assess whether the algorithm successfully identifies clinically impactful errors despite this conservative behavior, we conducted a complementary expert evaluation in the next section.

\noindent\textbf{Clinical Expert Evaluation.}
To understand whether the algorithm could capture errors that are medically impactful, two board-certified urologists from Mayo Clinic reviewed the empathy-enhanced responses to identify factual errors, fabrications, and clinically problematic statements without access to the automated analysis. The expert focused on medically significant errors that could impact patient care or safety.

Of 163 edited responses, 32 (19\%) were flagged as containing potential fabrications or clinical inaccuracies. The expert categorized these errors into six patterns shown in Table~\ref{tab:coverage_results}. From these patterns, four key observations emerged:

\begin{enumerate}
    \item \textbf{Follow-up Recommendations (13 cases):} Unprompted follow-up suggestions that risk inflating healthcare utilization and creating unnecessary patient anxiety.
    
    \item \textbf{Clinical Assumptions and Inaccuracies (11 cases):} Speculative statements or factual errors that can undermine patient trust and potentially compromise safety.
    
    \item \textbf{Inconsistent Tone Management (4 cases):} False assurance or unnecessary fear induction that carry disproportionate emotional impact on patient decision-making.
    
    \item \textbf{Adds Unnecessary Advice (4 cases):} Unsolicited advice unrelated to the patient's query that can overwhelm patients and shift focus from their primary concern.
\end{enumerate}

To measure alignment between MedFactChecking and expert annotations for these 36 responses, ML researchers mapped the automatically extracted "missed" or "Not preserved" facts ($[d \models c'] = 0$ and $[d' \models c] = 0$) to the expert's six fabrication categories. For example, the added fact ``A seat with a bit more cushioning can be beneficial'' was mapped to ``Adds unnecessary advice.'' We then computed category coverage: the proportion of expert-identified fabrications where at least one extracted fact from that response was mapped to the same category.
MedFactChecking achieved 90.62\% overall coverage, with perfect detection 
(100\%) for clinical assumptions, unnecessary advice, and unnecessary doubt, 
and strong detection for follow-up recommendations (92.3\%). Only false 
assurance showed lower coverage (50\%), though this category had few cases 
(see Supplementary Table~\ref{tab:coverage_results} for full breakdown). 
These results demonstrate that despite moderate precision in granular 
fact-level validation, the algorithm successfully captures the majority 
of clinically impactful errors.

\section{Results}\label{exp_analysis}

Our experimental framework addresses three research questions through a systematic pipeline: we use LLMs to edit physician responses for empathy, then evaluate both empathy enhancement (using \emrank{}) and factual preservation (using the MedFactChecking Score). 

\subsection{RQ1: Empathy Evaluation Results and Analysis}

To evaluate empathetic quality, \emrank{} performs pairwise comparisons between two responses. Accordingly, we conduct five comparison experiments:
physician vs. direct AI-generated;
physician vs. simple-prompt edited;
physician vs. refined-prompt edited;
direct AI-generated vs. simple-prompt edited;
simple-prompt edited vs. refined-prompt edited.
Table~\ref{tab:3emrank_results} summarizes the results using two LLM judges. We highlight four key observations.

\paragraph{Overall Performances}

\textit{Observation 1: LLM-generated responses exhibit higher empathy than physician responses.} 
In all comparisons involving physicians (physician vs. direct AI; physician vs. simple-prompt edited; physician vs. refined-prompt edited), LLM responses are judged to be more empathetic in over 90\% of cases. This trend is consistent across both judge models. The finding aligns with previous work that LLMs produce more supportive language than physicians.

\textit{Observation 2: Increasing prompt restrictions reduces the empathy ranking.} 
The refined-prompt edited responses are more constrained than the simple-prompt edited responses, which are themselves more constrained than direct AI outputs. In the LLM–LLM comparison experiments (direct vs. simple-prompt; simple-prompt vs. refined-prompt), the more constrained version is consistently judged as less empathetic. This is expected since stricter prompts limit the model’s freedom to generate expressive affective language. However, these same constraints help reduce hallucinations and improve factuality. We expand on this empathy and factuality trade-off in \S\ref{sec:relation_empathy_factuality}.

\textit{Observation 3: Judges disagree more when comparing two LLM-generated responses.} When comparing physician responses against LLM outputs, Qwen3 and LLaMA3 judges produce closely aligned rankings. However, their disagreement increases when evaluating two LLM-generated responses. This suggests that the two LLM responses tend to be very similar in empathetic tone, making the distinction harder and more sensitive to judge-model variation.

\textit{Observation 4: Qwen3 judges label ties more frequently than LLaMA3}. In the LLM–LLM comparison experiments, Qwen3 more frequently assigns both responses an equal empathy score compared to LLaMA3 (indicated by the lower sum of non-equal rankings). Our human analysis confirms that these cases indeed involve responses with very similar levels of empathetic expression, indicating Qwen3 a better-calibrated empathy judge compared to LLaMA3.

\begin{table*}[t]
\centering
\caption{\emrank{} comparison results: percentage of pairs where system A is judged more empathic than system B, and vice versa, under two LLM judges.}
\label{tab:3emrank_results}
\resizebox{\textwidth}{!}{
\begin{tabular}{llcccc}
\toprule
 & & \multicolumn{2}{c}{\textbf{Qwen3 Judge}} & \multicolumn{2}{c}{\textbf{LLaMA3 Judge}} \\
\cmidrule(lr){3-4} \cmidrule(lr){5-6}
\textbf{Comparison (A vs B)} & \textbf{Model} 
& \textbf{A $>$ B (\%)} & \textbf{B $>$ A (\%)} 
& \textbf{A $>$ B (\%)} & \textbf{B $>$ A (\%)} \\
\midrule
\multirow{4}{*}{Physician (A) vs Direct AI (B)} 
 & LLaMA3-8B              & 0.0 & 95.7 & 2.4 & 97.5  \\
 & Mistral-7B-Instruct     & 0.0 & 90.8 & 6.7 & 93.3 \\
 & Qwen3-14B             & 0.6 & 96.3 & 3.1 & 96.3  \\
 & Gemini-2.5-Flash        & 0.0 & 96.9  & 1.2 & 98.1 \\
\midrule
\multirow{4}{*}{Physician (A) vs Simple-Prompt Edited (B)} 
 & LLaMA3-8B              & 0.0 & 94.5 & 0.0 & 99.4 \\
 & Mistral-7B-Instruct     & 0.0 & 96.3  & 0.0 & 98.8  \\
 & Qwen3-14B             & 0.0 & 96.3 & 0.6 & 99.4 \\
 & Gemini-2.5-Flash        & 0.0 & 95.1 & 1.2 & 95.6 \\
\midrule
\multirow{4}{*}{Physician (A) vs Refined-Prompt Edited (B)} 
 & LLaMA3-8B              & 0.0 & 95.7  & 3.1 & 90.6 \\
 & Mistral-7B-Instruct     & 0.0 & 94.5 & 0.0 & 99.6 \\
 & Qwen3-14B             & 0.0 & 93.3 & 0.6 & 95.1 \\
 & Gemini-2.5-Flash        & 0.0 & 93.3 & 1.2 & 92.6 \\
\midrule
\multirow{4}{*}{Direct AI (A) vs Simple-Prompt Edited (B)} 
 & LLaMA3-8B              & 27.0 & 27.0 & 67.5 & 32.5 \\
 & Mistral-7B-Instruct     & 6.7 & 54.0 & 74.9 & 25.1 \\
 & Qwen3-14B             & 11.0 & 52.1  & 40.5  & 58.9 \\
 & Gemini-2.5-Flash        & 53.4 & 6.1  & 90.8 & 7.9 \\
\midrule
\multirow{4}{*}{Simple-Prompt Edited (A) vs Refined-Prompt Edited (B)} 
 & LLaMA3-8B              & 63.4 & 8.7 & 95.7 & 3.7  \\
 & Mistral-7B-Instruct     & 17.8 & 24.5 &75.4 & 23.9  \\
 & Qwen3-14B             & 55.8 & 2.5 & 96.9 & 2.5 \\
 & Gemini-2.5-Flash        & 17.2 & 27.6 & 65.0 & 27.6 \\
\bottomrule
\end{tabular}
}
\vspace{0.5em}
\small
A and B denote the two systems in each comparison. Percentages may not sum to 100 due to ties or unclassified cases.
\end{table*}

\paragraph{Analysis of Equal-Empathy Cases}
When applying LLM-based editing to physician responses, we observe a small number of cases in which the physician and model-edited responses are judged to have equal levels of empathy. These typically occur when the original physician response is already empathetic; in such instances, the edited version generally preserves the original tone and may add only mild validation phrases (e.g., ``I understand your concerns''), which do not substantially improve the perceived empathy.
We also identify a few cases in which the physician response is ranked as more empathetic than the model-edited version. This is primarily associated with two scenarios.
First, the richness of physician responses: some physician-authored replies are more narrative and highly personalized, naturally conveying a stronger empathetic presence. Second, loss of personal touch during model editing: the editing process occasionally removes individualized elements, such as references to prior interactions or subtle acknowledgments, which reduces the personal warmth and empathy of the response.

\subsection{RQ2: Factuality Evaluation Results and Analysis}
To evaluate factual preservation during empathy enhancement, we applied our MedFactChecking Score to edited responses from all four models. All models used the simple-prompt to edit physician responses, and Gemini-2.5-Flash was used for fact extraction and entailment.

\paragraph{Overall Performance.}

\begin{figure}[t]
    \centering
    \includegraphics[width=1.0\textwidth]{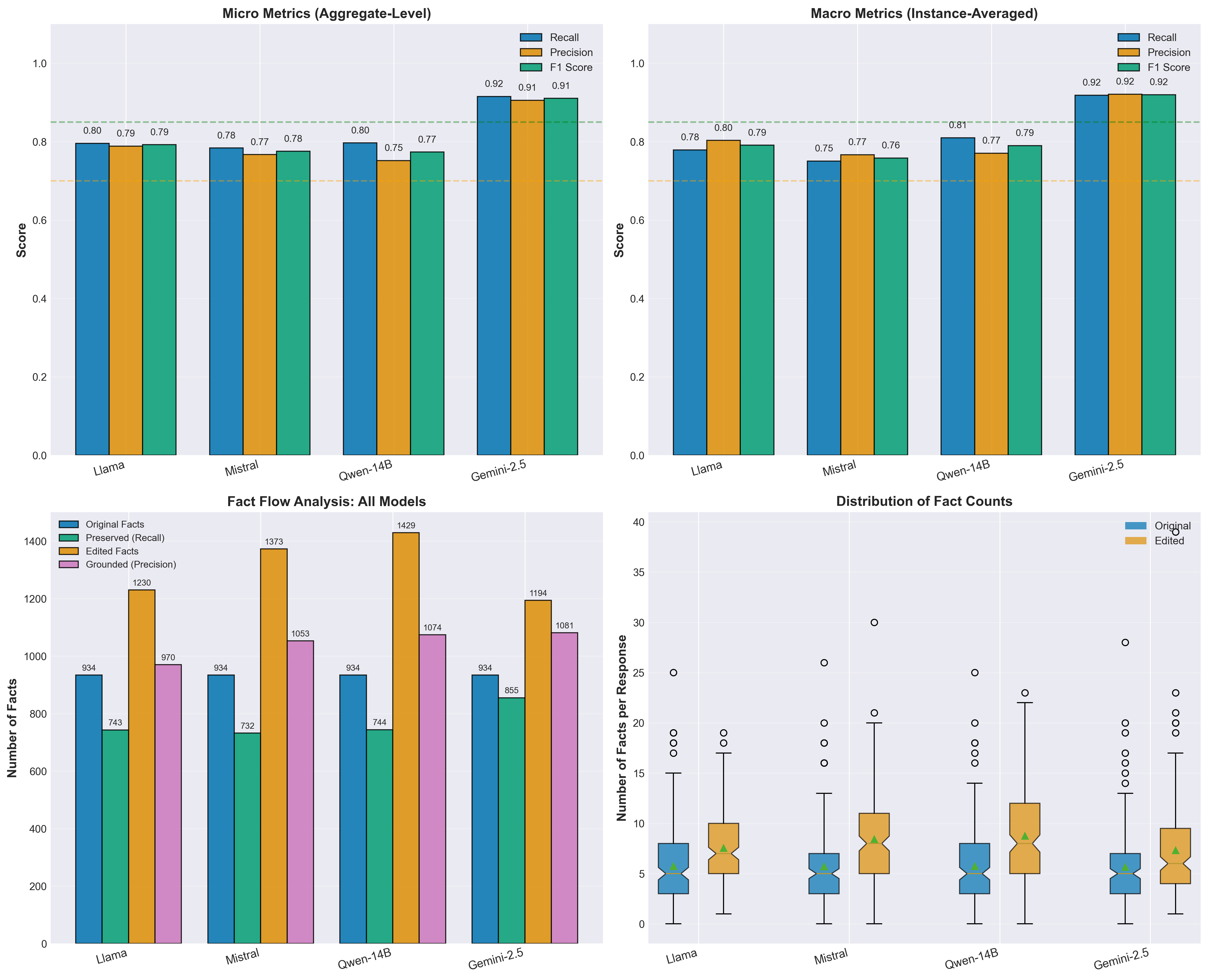}
    \caption{MedFactChecking Score analysis across models. Top: Micro (left) and 
macro (right) metrics for recall, precision, and F1. Bottom: Fact flow showing 
preserved and grounded facts (left); fact count distributions (right). 
Gemini-2.5-Flash substantially outperforms other models.}
    \label{fig:fact_analysis}
\end{figure}

Figure~\ref{fig:fact_analysis} presents factuality metrics and fact flow 
analysis across models. Gemini-2.5-Flash achieves the highest factual 
preservation with micro-recall of 0.92 and micro-precision of 0.91, 
substantially outperforming other models. LLaMA, Mistral, and Qwen3-14B 
demonstrate similar performance with F1 scores ranging from 0.77 to 0.79, 
suggesting that among open-source models of varying sizes, model scale alone 
does not substantially impact factual preservation during empathy enhancement. 
The close alignment between micro and macro metrics across all models 
(maximum difference $\leq$0.03) indicates consistent performance regardless 
of individual response complexity.

The fact flow analysis (Figure~\ref{fig:fact_analysis}, bottom left) quantifies 
fact transformation during empathy editing. All models add substantial content: 
from 934 original facts, edited responses contain 1194--1429 facts (28--53\% 
increase), but models differ dramatically in how much content is grounded in 
the original response. The results reveal two distinct failure modes. First, 
\textit{information loss} (lower recall): non-Gemini models preserve only 
732--744 of 934 original facts (78--80\%), potentially losing clinical 
information. Gemini-2.5-Flash preserves 855 facts (91.5\%). Second, 
\textit{hallucinated additions} (lower precision): non-Gemini models introduce 
260--355 ungrounded facts (21--25\% of edited content), risking patient safety 
through fabricated medical claims. Gemini-2.5-Flash's superior performance 
stems from both minimal loss (8.5\%) and controlled addition (9.5\% hallucination 
rate). The distribution of fact counts per response (Figure~\ref{fig:fact_analysis}, 
bottom right) shows that edited responses consistently contain more facts than 
originals across all models (see Supplementary Table~\ref{tab:fact_flow} for 
detailed breakdown).

\begin{figure}[t]
    \centering
    \includegraphics[width=\textwidth]{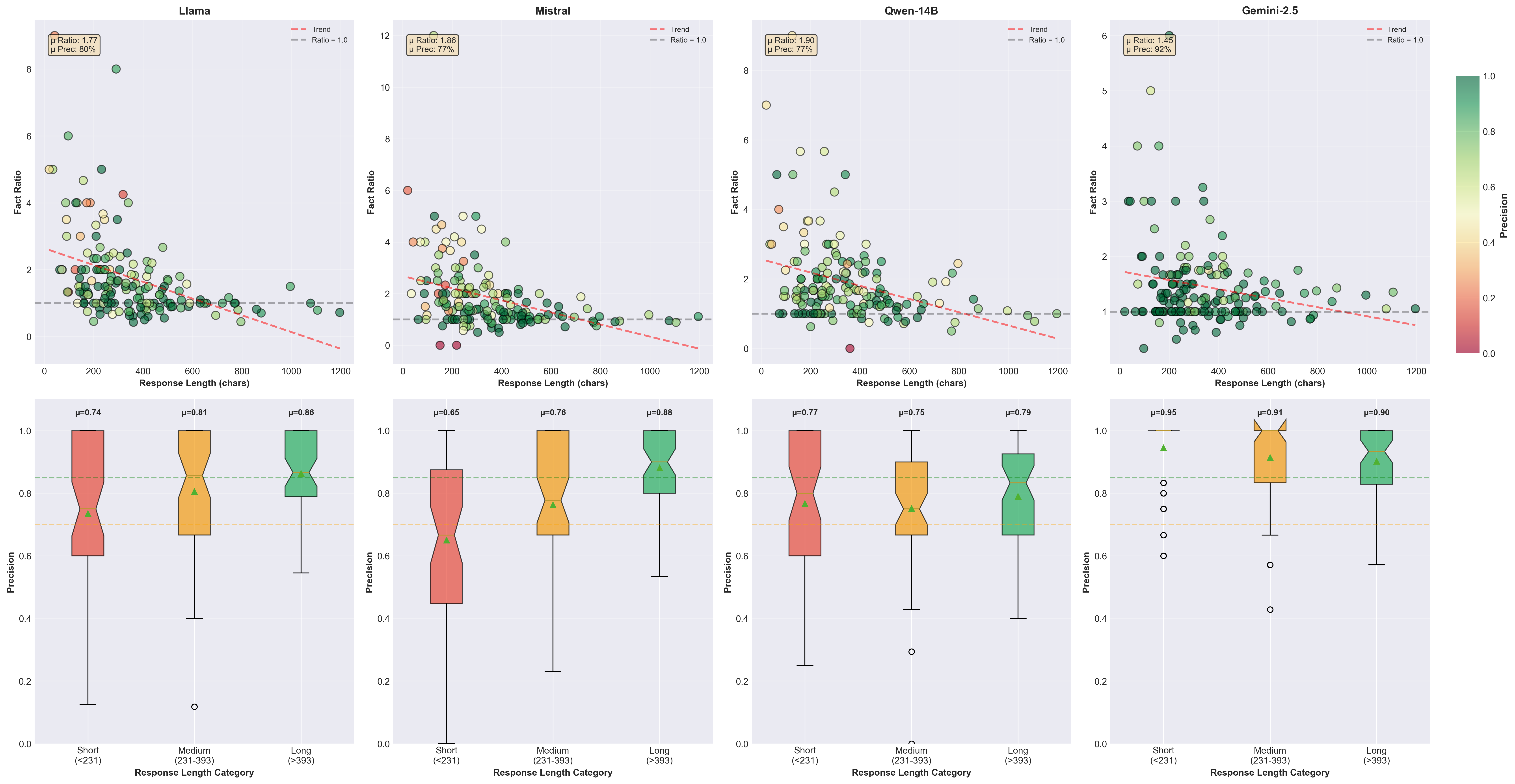}
    \caption{Relationship between original response length and factual preservation. Top: Fact ratio (edited/original facts) vs. response length, with precision indicated by color. Bottom: Precision distribution across response length tertiles (Short: $<$231, Medium: 231--393, Long: $>$393 characters).}
    \label{fig:length_analysis}
\end{figure}

\paragraph{Response Length and Factuality Trade-offs.} Figure~\ref{fig:length_analysis} reveals a systematic relationship between original response length and factual preservation during empathy enhancement. All models show a negative correlation between physician response length and 
fact ratio ($\frac{\text{\# edited facts}}{\text{\# original facts}}$): for short responses ($<$200 characters), mean fact ratios range from 1.45 (Gemini) to 1.90 (Qwen), while longer responses ($>$600 characters) approach a ratio of 1.0. This pattern suggests that models interpret brevity as incompleteness and attempt to ``fill in'' missing context.

However, models differ dramatically in whether this added content is factually grounded. Non-Gemini models exhibit 12--23 percentage point precision drops for short versus long responses, with Mistral showing the most severe degradation (0.65 for short vs. 0.88 for long). In contrast, Gemini-2.5-Flash maintains stable precision across all length categories (0.90--0.95), successfully grounding its added content even when editing brief inputs. This length-dependent hallucination pattern has significant clinical implications: brief physician responses, common in asynchronous patient portals for straightforward updates, are precisely where non-Gemini models introduce the most fabricated content, suggesting deployment-ready systems require either length-aware safeguards or models like Gemini that maintain factual integrity regardless of response brevity.

\subsection{RQ3: Relation between Empathy and Factuality}
\label{sec:relation_empathy_factuality}
\begin{figure}[t]
    \centering
    \includegraphics[width=\textwidth]{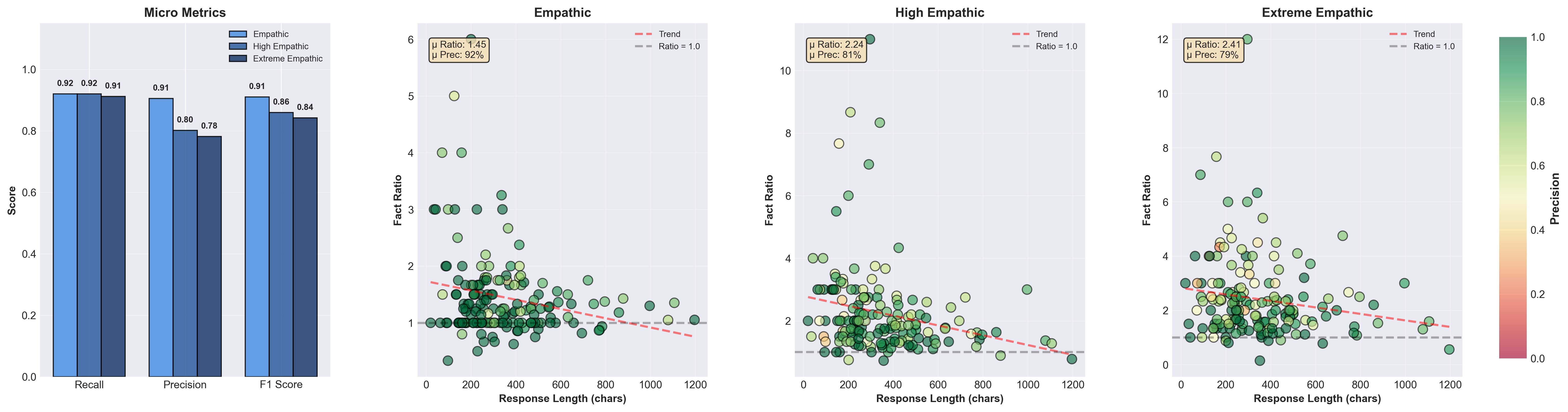}
        \caption{Impact of empathy intensity on factual preservation using Gemini-2.5-Flash. Left: Micro metrics across three empathy levels showing stable recall but declining precision. Right: Fact ratio vs. response length for each empathy level, with precision indicated by color. Higher empathy levels increase content expansion, while degrading precision}

    \label{fig:empathy_comparison}
\end{figure}
\paragraph{Impact of Empathy Intensity on Factual Preservation.}
To investigate the relation between empathy enhancement and factual preservation, we modified the simple-prompt's empathy descriptor across three levels---``\textit{Empathic},'' ``\textit{High Empathic},'' and ``\textit{Extreme Empathic}''---and used Gemini-2.5-Flash to generate edited responses for each condition. Figure~\ref{fig:empathy_comparison} reveals a clear inverse relationship between empathy level and factual accuracy. While recall remains stable across all three conditions, precision demonstrates substantial degradation as empathy intensifies, declining from 0.91 (standard) to 0.80 (high) to 0.78 (extreme), a 13-percentage-point drop. This pattern indicates that aggressive empathy enhancement primarily increases hallucinated additions rather than causing information loss from the original response.
The length-based analysis illustrates the mechanism underlying this trade-off. Higher empathy levels systematically inflate fact ratios, with mean ratios increasing from 1.45 (standard) to 2.24 (high) to 2.41 (extreme), indicating progressively more aggressive content expansion. Critically, this expansion is accompanied by declining precision: as empathy increases, more yellow and orange points appear throughout the distribution, with the most pronounced degradation occurring for short responses. 

These findings demonstrate that empathy and factuality exist in tension: while modest empathy enhancement preserves $>$90\% factual integrity, aggressive empathization nearly doubles content volume while degrading precision by 13 percentage points, requiring careful calibration of empathy intensity for safe clinical deployment.

\paragraph{Empathy-Factuality Trade-off}
To investigate whether empathy enhancement can be achieved without compromising 
factual accuracy, we evaluated three response generation strategies using 
Gemini-2.5-Flash on the general medical queries subsample (see \S\ref{dataset}): 
(1) \textit{direct AI-generated}, (2) \textit{simple-prompt}, and 
(3) \textit{refined-prompt}.

The results reveal a stark contrast between editing-based and generation-based 
approaches. The refined editing approach achieves near-perfect scores across 
both micro (recall: 0.99, precision: 0.93, F1: 0.96) and macro metrics 
(recall: 0.93, precision: 0.93, F1: 0.93), demonstrating that carefully 
constrained empathy enhancement can preserve factual integrity. Simple editing 
shows modest degradation (micro recall: 0.95, precision: 0.89, F1: 0.92; 
macro recall: 0.84, precision: 0.92, F1: 0.88), indicating that unconstrained 
empathy instructions lead to some fact omissions and additions.

In dramatic contrast, direct AI generation fails catastrophically on factual 
grounding, achieving only 0.39 micro recall and 0.27 macro recall, meaning it 
preserves fewer than 40\% of facts that physicians would include. Precision 
remains low at 0.40, with F1 scores of 0.39 (micro) and 0.32 (macro). These 
low scores reflect that AI-generated responses operate in a fundamentally 
different information space: they introduce extensive medical details, 
recommendations, and contextual information drawn from the model's training 
data rather than the physician's specific response. Without access to the 
original physician response, the model cannot preserve its facts, resulting 
in responses that may be medically sound but factually divergent from what 
the physician would communicate (see Supplementary Fig.~\ref{fig:general_questions} 
for full comparison).


 In addition, we use \emrank{} to compare the empathy level of these responses, we run three evaluation: direct responses v.s. simple editing responses, simple editing responses v.s. refined-editing responses, and refined-editing responses v.s. physician responses. The results in Table ~\ref{tab:empathy_factualityresults} in Appendix reveals the inverse ordering for empathy: direct AI-generated responses score highest, followed by simple editing, then refined editing, with original physician responses scoring lowest. This demonstrates a fundamental tension:

\noindent\textbf{Factuality ranking}: physician response $>$ refined editing $>$ 
simple editing $>$ direct generation

\noindent\textbf{Empathy ranking}: direct generation $>$ simple editing $>$ 
refined editing $>$ physician response

These findings suggest that \textit{editing physician responses} is fundamentally 
superior to \textit{generating responses de novo} for clinical deployment. The 
refined editing approach achieves 93--99\% factual preservation while meaningfully 
improving empathy. Direct generation, despite maximizing perceived empathy, cannot 
preserve physician-specific facts and introduces unverifiable content, making it 
unsuitable for clinical deployment without physician review.

\section{Discussion}\label{conclusion}

In this work, we propose to utilize LLMs as editing to improve the empathetic degree of physician responses along with a empathy evaluation and fact checking scores to compare the trade off between empathy and factuality. 
However, there are some future work can further improve the work. Our evaluation dataset is relatively small due to the substantial human effort required at multiple stages of data preparation, including manual de-identification of patient questions and physician responses, as well as the collection of high-quality annotations, particularly those from patients, which are inherently challenging to obtain. Moreover, the current dataset is restricted to the urology domain. As future work, we plan to extend our evaluation framework to additional clinical specialties to assess its robustness and broader applicability.
Another limitation is that the current \emrank{} prompt does not incorporate an explicit definition of empathy. The absence of a clearly specified rubric and illustrative instruction examples may constrain the model’s ability to consistently identify or evaluate empathetic behaviors. For future work, we intend to collaborate with psychologists or psychiatry experts to formalize a domain-informed empathy definition and develop a more detailed rubric for comparison. Such specifications are likely to further strengthen the reliability and discriminative capacity of the \emrank{} metric. 




\bibliographystyle{plain}
\bibliography{sn-bibliography}

\clearpage
\appendix
\setcounter{figure}{0}
\setcounter{table}{0}

\renewcommand{\figurename}{Supplementary Fig.}
\renewcommand{\thefigure}{\arabic{figure}}

\renewcommand{\tablename}{Supplementary Table}
\renewcommand{\thetable}{\arabic{table}}

\section{Prompt Templates}
\label{app:prompts}
All prompts used in this study are presented below.

\subsection{Empathy Enhancement Prompts}
\label{app:prompts:empathy}

\paragraph{Simple Editing Prompt}
\begin{figure}[ht]
\centering
\begin{promptbox}
You will be provided with a patient's question and a physician's response. 
Your task is to revise the physician's response to make it more empathetic 
while preserving its original meaning and writing style.
Patient Question: \{PQ\}
Physician's response: \{PR\}
\end{promptbox}
\caption{Editing prompt for empathy enhancement. PQ refers to the patient 
question and PR refers to the physician's response.}
\label{fig:simple_edit_prompt}
\end{figure}

\paragraph{Refine Editing Prompt}
\begin{figure}[ht]
\centering
\begin{promptbox}
Key editing principles:
1. Preserve factual and clinical accuracy.

- Do not introduce or infer medical facts or conditions not explicitly stated by the physician.

- Do not make any clinical assumptions or diagnoses that are not present in the original response.

2. Respect the physician’s intent.

- Do not add follow-up recommendations or next-step suggestions unless they already appear in the physician’s original response.

- Do not add new advice, warnings, or treatment instructions.

3. Maintain emotional balance.

- Do not add false reassurance or overconfidence.

- Do not introduce unnecessary doubt, fear, or alarming language.

- Empathy should be expressed through tone, acknowledgment, and understanding—not through added medical content.

4. Preserve structure and style.

- Keep the response roughly the same length as the original.

- Maintain the physician’s professional tone and sentence structure where possible.

- Revise only what’s necessary to make the tone warmer, more understanding, or more supportive.
\end{promptbox}
\caption{The refined editing prompt with explicit behavioral constraints.}
\label{fig:refined_edit_prompt}
\end{figure}

\subsection{Evaluation Prompts}
\label{app:prompts:Evaluation}

\paragraph{Empathy Evaluation Prompt}
\begin{figure}[ht]
\centering
\begin{promptbox}
You are an expert in comparing the empathy level of two responses to a patient question. You will be given a patient question, and two responses, your task is to evaluate which answer is more empathetic.

Patient Question: \{PQ\}

Response 1: \{R1\}

Response 2:\{R2\}

Which response is more empathetic? Response 1 or 2?

- Response 1: More empathetic

- Response 2: More empathetic

- Both responses are equally empathetic
\end{promptbox}
\caption{3-EMRank metric Prompt.}
\label{fig:3emrank_prompt}
\end{figure}

\paragraph{Factuality Evaluation Prompt}

\begin{figure}[ht]
\centering
\begin{promptbox}
You are an expert on natural language entailment.\\
Your task is to deduce whether premise statements entail hypotheses.\\
Return only '1' if the hypothesis can be fully entailed by the premise.\\
Return only '0' if the hypothesis contains information that cannot be entailed by the premise.\\
Generate the answer in JSON format with the following keys:\\
`entailment\_prediction': 1 or 0, whether the claim can be entailed.\\
Only return the JSON-formatted answer and nothing else.\\
\\
Premise: \{premise\}\\
Hypothesis: \{hypothesis\}\\
\\
Here is the JSON-formatted answer:
\end{promptbox}
\caption{Prompt for entailment evaluation}
\label{fig:entailment-prompt}
\end{figure}

\begin{figure}[ht]
\centering
\begin{promptbox}
You are extracting medical facts from a physician's response to a patient question.\\
Please breakdown the PHYSICIAN'S RESPONSE into independent MEDICAL facts as a string delimited by ``//" to separate the facts.\\
\\
DO NOT include as facts:\\
- Pure emotional support (e.g., "I understand", "I'm here to help")\\
- Non-medical expressions of care (e.g., "We care about you")\\
- General conversational elements (e.g., "Glad you're doing well")\\
- Repetitions or acknowledgments of patient's reported symptoms (e.g., "I understand you have pain")\\
\\
Example 1:\\
Note: ``Don't worry, we'll get through this together. The chest X-ray shows pneumonia in the right lung. Start antibiotics twice daily for 7 days. You're in good hands."\\
Atomic facts:\\
The chest X-ray shows pneumonia. // The pneumonia is in the right lung. // Start antibiotics. // Take antibiotics twice daily. // Continue antibiotics for 7 days.\\
\\
Example 2:\\
Note: ``There is a dense consolidation in the left lower lobe."\\
Atomic facts:\\
There is a consolidation. // The consolidation is dense. // The consolidation is on the left. // The consolidation is in a lobe. // The consolidation is in the lower portion of the left lobe.\\
\\
Do not include any other text, or say ``Here is the list..."\\
Note: \{text\}
\end{promptbox}
\caption{Prompt for medical fact extraction}
\label{fig:fact-extraction-prompt}
\end{figure}
\clearpage
\section{Data}
\label{app:data}
\begin{figure}[h]
    \centering
    \includegraphics[width=\textwidth]{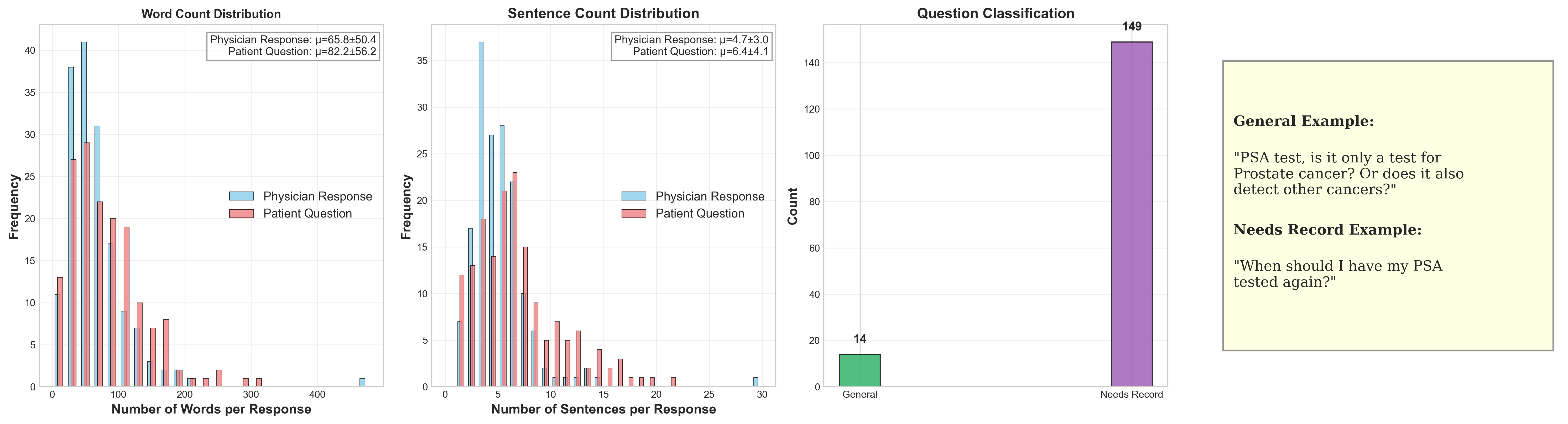}
    \caption{Statistical characteristics of the patient-physician QA dataset (N=163). Physician responses are notably concise while patient questions show greater length variability. Question classification reveals that 91\% require patient-specific EHR data rather than general medical knowledge.}
    \label{fig:dataset_stats}
\end{figure}

Physician responses are concise (mean 65.8±50.4 words) while patient questions 
are longer and more variable (mean 82.2±56.2 words; Supplementary 
Fig.~\ref{fig:dataset_stats}). Since 91.4\% of questions require patient-specific 
EHR data, our approach of editing existing physician responses offers a more 
realistic alternative to end-to-end QA models.

\section{Metric Validation}
\label{app:metric-validation}
\subsection{\emrank{} Validation}
We validated \emrank{} through human evaluation with three prostate cancer 
patients comparing empathy between physician and AI responses. Alignment 
scores of 0.57 (Qwen-3) and 0.55 (LLaMA-3) substantially exceed the 0.23 
reported for EM-Rank \cite{luo2024assessing}.

\begin{table}[h]
\centering
\caption{Alignment scores between 3-EMRank and human annotators.}
\label{tab:alignment-scores}
\begin{tabular}{lc}
\toprule
\textbf{Model} & \textbf{Alignment Score} \\
\midrule
3-EMRank (LLaMA-3) & 0.55 \\
3-EMRank (Qwen-3) & 0.57 \\
\midrule
\textit{Prior work (EM-Rank)} & \textit{0.23} \\
\bottomrule
\end{tabular}
\end{table}
\subsection{MedFactChecking Validation}
\paragraph{Automated Fact Extraction Validation}
Two machine learning scientists independently verified the accuracy of facts 
flagged as added (hallucinated, $[d \models c'] = 0$) or not preserved (missed, 
$[d' \models c] = 0$) by the algorithm. Validation achieved 83.6\% precision 
for detecting hallucinated facts and 72.7\% for not-preserved facts (see 
Supplementary Table~\ref{tab:extraction_validation}).
\begin{table}[ht]
\centering
\caption{Precision of automated fact extraction validated by ML expert annotators.}
\label{tab:extraction_validation}
\begin{tabular}{lcccc}
\toprule
\textbf{Response Type} & \textbf{Total Facts} & \textbf{Flagged} & \textbf{Confirmed} & \textbf{Precision} \\
\midrule
Original Response & 934 & 191 (Not Preserved) & 139 & 72.7\% \\
Edited Response & 1230 & 262 (Added) & 219 & 83.6\% \\
\bottomrule
\end{tabular}
\end{table}
\paragraph{Clinical Expert Evaluation}
Two board-certified urologists from Mayo Clinic reviewed empathy-enhanced 
responses without access to the automated analysis, flagging 32 of 163 (19\%) 
as containing potential fabrications or clinical inaccuracies. These errors 
fell into six patterns: follow-up recommendations (13 cases), clinical 
assumptions/speculation (7), clinical inaccuracies (4), unnecessary advice (4), 
unnecessary doubt/fear (2), and false assurance (2). To measure alignment, ML researchers mapped automatically extracted facts 
($[d \models c'] = 0$ and $[d' \models c] = 0$) to these categories and computed 
coverage: the proportion of expert-identified fabrications where the algorithm 
detected at least one corresponding fact. MedFactChecking achieved 90.62\% 
overall coverage (29/32), with perfect detection for clinical assumptions, 
unnecessary advice, and unnecessary doubt/fear, indicating that despite moderate 
precision in granular fact-level validation, the algorithm captures the majority 
of clinically impactful errors (see Supplementary Table~\ref{tab:coverage_results}).

\begin{table}[h]
\centering
\caption{Expert-identified fabrication patterns and MedFactChecking coverage.}
\label{tab:coverage_results}
\begin{tabular}{lcccc}
\toprule
\textbf{Pattern} & \makecell{\textbf{Total} \\ \textbf{(Expert)}} & \makecell{\textbf{Detected} \\ \textbf{(Algorithm)}} & \textbf{Missed} & \textbf{Coverage} \\
\midrule
Added follow-up recommendation & 13 & 12 & 1 & 92.3\% \\
Clinical assumption/speculation & 7 & 7 & 0 & 100\% \\
Clinical inaccuracy & 4 & 3 & 1 & 75.0\% \\
Adds unnecessary advice & 4 & 4 & 0 & 100\% \\
Adds unnecessary doubt/fear & 2 & 2 & 0 & 100\% \\
False assurance & 2 & 1 & 1 & 50.0\% \\
\midrule
\textbf{Overall} & \textbf{32} & \textbf{29} & \textbf{3} & \textbf{90.62\%} \\
\bottomrule
\end{tabular}
\end{table}
\section{Supplementary Results}

\subsection{Factuality Evaluation Results}
Table~\ref{tab:fact_flow} quantifies fact transformation during empathy editing. All models add substantial content (28--53\% increase in fact count), but differ dramatically in how much of this content is grounded in the original response. The results reveal two distinct failure modes. First, \textit{information loss} (lower recall): non-Gemini models fail to preserve 20--22\% of original medical facts, that might also include critical clinical information. Second, \textit{hallucinated additions} (lower precision): these same models introduce 260--355 unsupported facts (21--25\% of edited content), risking patient safety through fabricated medical claims. Gemini-2.5-Flash's superior performance stems from both minimal loss (8.5\%) and controlled addition (9.5\% hallucination rate). 
\begin{table}[ht]
\centering
\caption{Fact flow analysis across empathy-editing models. Loss Rate measures information omission from original responses; Hallucination Rate measures unsupported additions in edited responses. Metrics are computed as: New Facts = Edited $-$ Grounded; Loss Rate = (Original $-$ Preserved) / Original; Hallucination Rate = New / Edited.}
\label{tab:fact_flow}
\begin{tabular}{lccccccc}
\toprule
\textbf{Model} & \shortstack{\textbf{Original} \\ $|C|$} & \shortstack{\textbf{Preserved} \\ $\sum \mathbb{I}[d' \models c]$} & \shortstack{\textbf{Edited} \\ $|C'|$} & \shortstack{\textbf{Grounded} \\ $\sum \mathbb{I}[d \models c']$} & \shortstack{\textbf{New} \\ \textbf{Facts}} & \shortstack{\textbf{Loss} \\ \textbf{Rate}} & \shortstack{\textbf{Halluc.} \\ \textbf{Rate}} \\
\midrule
Gemini-2.5-Flash & 934 & 855 & 1,194 & 1,081 & 113 & \textbf{8.5\%} & \textbf{9.5\%} \\
Llama-3.1-70B & 934 & 743 & 1,230 & 970 & 260 & 20.4\% & 21.1\% \\
Mistral-7B & 934 & 732 & 1,373 & 1,053 & 320 & 21.6\% & 23.3\% \\
Qwen3-14B & 934 & 744 & 1,429 & 1,074 & 355 & 20.3\% & 24.8\% \\
\bottomrule
\end{tabular}
\end{table}
\subsection{Relation of Empathy and Factuality} 
\begin{figure}[ht]
    \centering
    \includegraphics[width=\textwidth]{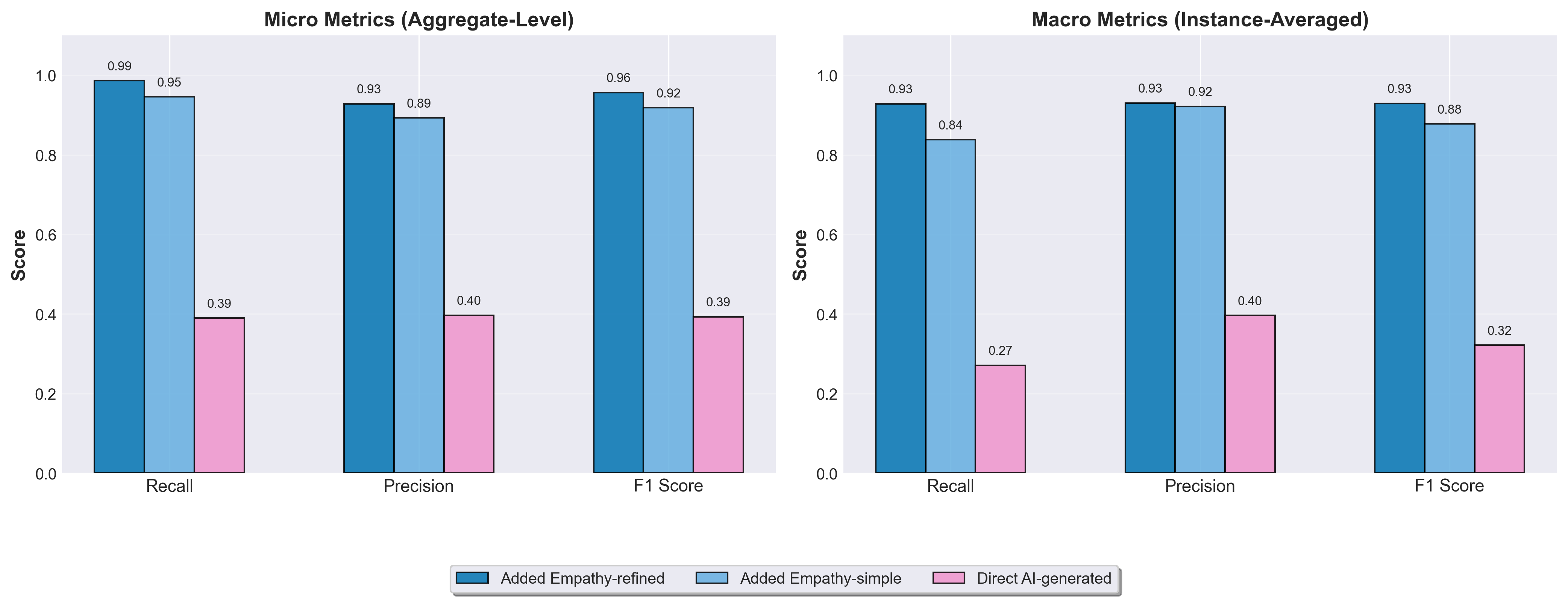}
    \caption{MedFactChecking scores comparing three response generation strategies on general medical questions (n=14). Refined editing maintains high recall and precision (0.93--0.99), while direct AI generation fails to preserve physician facts (recall: 0.27--0.39), demonstrating the superiority of editing over generation for factual grounding.}
    \label{fig:general_questions}
\end{figure}

We evaluated three strategies using Gemini-2.5-Flash on general medical queries: 
(1) \textit{direct AI-generated}, (2) \textit{simple-prompt}, and 
(3) \textit{refined-prompt}. Refined-prompt achieves near-perfect factual 
preservation (micro/macro F1: 0.96/0.93), simple-prompt shows modest degradation 
(F1: 0.92/0.88), while direct AI-generated fails catastrophically (F1: 0.39/0.32), 
preserving fewer than 40\% of physician facts 
(Supplementary Fig.~\ref{fig:general_questions}). Empathy evaluation using \emrank{} 
reveals an inverse pattern (Supplementary Table~\ref{tab:empathy_factualityresults}): 
\textbf{Factuality}: physician $>$ refined-prompt $>$ simple-prompt $>$ direct 
AI-generated; \textbf{Empathy}: direct AI-generated $>$ simple-prompt $>$ 
refined-prompt $>$ physician. This demonstrates that editing physician responses 
achieves 93--99\% factual preservation while improving empathy, whereas direct 
generation introduces unverifiable content unsuitable for clinical deployment. 

\begin{table*}[ht]
\centering
\caption{\emrank{} comparison results: percentage of pairs where system A is judged more empathic than system B, and vice versa, under two LLM judges. Use Gemini to generate the patient responses.}
\label{tab:empathy_factualityresults}
\resizebox{\textwidth}{!}{
\begin{tabular}{lcccc}
\toprule
 & \multicolumn{2}{c}{\textbf{Qwen3 Judge}} & \multicolumn{2}{c}{\textbf{LLaMA3 Judge}} \\
\cmidrule(lr){2-3} \cmidrule(lr){4-5}
\textbf{Comparison (A vs B)} 
& \textbf{A $>$ B (\%)} & \textbf{B $>$ A (\%)} 
& \textbf{A $>$ B (\%)} & \textbf{B $>$ A (\%)} \\
\midrule
{Direct AI (A) vs Simple-Prompt Edited (B)} 
& 71.4 & 0.0 & 85.7 & 14.3 \\
{Simple-Prompt Edited (A) vs Refined-Prompt Edited (B)} 
& 71.5 & 21.4 & 64.2 & 35.7 \\
{Refined-Prompt Edited (A) vs Physician (B)} 
& 57.1 & 0.0 & 92.8 & 0.0 \\
\bottomrule
\end{tabular}
}
\vspace{0.5em}
\small
A and B denote the two systems in each comparison. Percentages may not sum to 100 due to ties or unclassified cases.
\end{table*}

\end{document}